\documentclass[a4paper]{article}
\usepackage[square,numbers]{natbib}
\usepackage[utf8]{inputenc}
\usepackage{pgf}
\usepackage{amsmath}
\usepackage{booktabs}
\usepackage{csquotes}
\usepackage{hyperref}
\providecommand{\keywords}[1]{\textbf{\textit{Keywords---}} #1}

\title{LLM Cognitive Judgements Differ From Human}
\author{Sotiris Lamprinidis \\ \href{mailto:sotiris@lamprinidis.com}{sotiris@lamprinidis.com}}
\date{Copenhagen, Denmark}

\begin{document}
\maketitle

\begin{abstract}
  Large Language Models (LLMs) have lately been on the spotlight of researchers,
  businesses, and consumers alike. While the linguistic capabilities of
  such models have been studied extensively, there is growing interest
  in investigating them as \emph{cognitive subjects}. In the present work I examine
  GPT-3 and ChatGPT capabilities on an limited-data inductive reasoning task 
  from the cognitive science literature. The results suggest that these models'
  cognitive judgements are not human-like.
\end{abstract}


\section{Introduction}
Large Language Models (LLMs), as typified by OpenAI
ChatGPT~\cite{chatgpt} have lately had massive impact on public opinion: out of a maximum of 100, the
terms ``ChatGPT'' and ``AI'' surged from 0 and 28 respectively on 20 Nov. 2022
to 73 and 99 respectively on 30 Apr.  2023~\footnote{Data source: Google Trends
(https://www.google.com/trends).} This optimism is also reflected in the
Markets, as valuations of ``AI''-exposed companies have been
surging~\cite{economist_2023_surging}. Researchers have expressed mixed 
opinions about the true abilities of such models, ranging from strong
claims about Artificial General Intelligence~\cite{bubeck2023sparks}(Microsoft
is a co-owner of OpenAI) and the seemingly anthropomorphic attribution of
consciousness on such models~\cite{lloyd2023like}
(which humans partially agree~\cite{scott2023you}), to dismissal of any
emergent notion of intelligence or human-like behavior~\cite{chomsky2023noam,floridi2023ai,katzir2023large}.

Perhaps most eloquently presented in~\cite{bender2021dangers}, it is not clear
if LLMs \emph{understand} language, or if they are just very proficient at
manipulating language symbols, which we humans are vulnerable to implicitly
interpreting as coherent. In fact, even before the current generation of LLMs
there has been a focus on uncovering general linguistic competencies and
capabilities
\cite{gulordava2018colorless,ribeiro2020beyond,warstadt2019neural,qian2019neural},
as well as the \emph{human-likedness} of Language
Models~\cite{gulordava2018colorless,ettinger2020bert}.
In the same spirit, there have been extensive investigations of ChatGPT
performance in Natural Language Processing (NLP)
tasks~\cite{bang2023multitask}, discovering good overall performance but poor
ability when it comes to reasoning and specifically inductive reasoning, as well
as extensively accounting ChatGPT failures~\cite{borji2023categorical}. Additionally,
ChatGPT responses seem to accurately mimic human language use~\cite{cai2023does}.
Could these positive results support claims of intelligence or consciousness of LLMs?

I propose that the strength of the evidence should be proportional 
to the strength of the claims made, and as such any emergent behaviours that seem
to transcend the mere linguistic realm when it comes to the capacity of LLMs should
be considered from a perspective of \emph{cognition}.

\section{Previous Work}

An early work on GPT-3~\cite{brown2020language} examined the responses qualitatively in what
is designated an ``Author's Turing Test'' and found inconsistent behaviour: 
``\emph{
[GPT-3 is] not better than our very best writers and philosophers at their peak. Are its best moments better than many humans and even perhaps, our best writers at their worst? Quite possibly}~\cite{elkins2020can}.
Other works have focused on evaluating LLMs in experiments from the cognitive
psychology literature and the \emph{human-likedness} of their responses.
\cite{binz2023using} found GPT-3 to mostly give human-like responses.
A recurring problem, however, with these studies is the explicit memorization
of the materials. Asking GPT-3 about the tasks in~\cite{binz2023using}, I found
that many of the tasks employed (all of Kahneman \& Tversky's, Wason's Card
Selection task, and the Blicket task) could be accurately recited by the
model.

An assortment of works study the Theory of Mind (ToM) of LLMs - the ability
to attribute mental states to others.  \cite{holterman2023does} examined both
ChatGPT and GPT-4 ToM using material from Kahneman \& Tversky. Similarly
to~\cite{binz2023using}, I found that ChatGPT could accurately recite
all the tasks.
The GPT-4 ToM was also qualitatively investigated in~\cite{freundexploring} using
10 philosophical paradoxes. Here too, I found that ChatGPT (the predecessor
model) could recite 9 out of the 10 paradoxes.  Another work focusing on
ChatGPT ToM applied clinical methods used to assess pathological ToM in humans
~\cite{brunet2023conversational}. The model in general performed poorly and not
similarly to humans in most tasks.

More closely related to the present work,~\cite{loconte2023challenging}
performed a neuropsychological investigation of abilities related to prefrontal
functioning and found ChatGPT unable to mimic human cognitive functioning
accurately. \cite{lipkin2023evaluating} studied GPT-3 as pragmatic reasoner and
found human-like behaviour, albeit with some exceptions. Finally,
\cite{xu2023does} investigated ChatGPT and GPT-4 on a range of psychological
tasks classified by the degree of embodiment associated with each and found
that LLM responses deviated from human on sensory and motor judgements.

In this work, I focus on~\cite{griffiths2006optimal} from the
cognitive science literature. The authors support the hypothesis that
when faced with inductive problems with limited data available,
human cognitive judgments closely correspond to those of a Bayesian
model with fitting priors. The motivation for choosing this particular
study is firstly there is no true answer to be determined deductively and as
such it seems ideal to reveal the underlying cognitive workings of a mind,
and secondly GPT-3 had no knowledge whatsoever about the work, while ChatGPT
was able to give a somewhat accurate high-level overview but could not recite
any particular experiment presented in the original work, thus
avoiding the common pitfall of model memorization of human responses.

\section{Methods}
In~\cite{griffiths2006optimal} all tasks involve predicting an extent
or duration of a phenomenon given an intermediate value at an unknown time:
given a value $t$, predict the final, total value $t_\text{total}$.
Responses from 350 participants were compared to a Bayesian predictor
which computes a probability over $t_\text{total}$:
${p(t_\text{total}|t)\propto p(t|t_\text{total})p(t_\text{total})}$.
The authors found that human judgments are close to the optimal predictions
of the Bayesian predictor.

Specifically, the tasks and respective $t$ are:
cake baking times (minutes), life spans (years), movie grosses (million US\$),
poem lengths (number of lines), U.S. representatives' terms (years), and
telephone box office waiting times (minutes).
I omit the ``Movie Runtimes'' and ``Pharaohs'' tasks that had high variance in
human responses. I inferred the values for the mean human predictions and
confidence intervals, as well as the predictions of the Bayesian models from
the plots in the original work. I refer to~\cite{griffiths2006optimal} for
further details on the prompts and the Bayesian predictors' choice of priors.

For each task and prompt value I asked GPT-3~\cite{gpt3model} and ChatGPT~\cite{chatgptmodel} the equivalent
question 20 times, starting a new conversation each time. I skipped the overall
introduction as it appears in the study and finished each prompt with
\emph{``The answer must be a single number, not a range, with no
explanation.''.} E.g.: \emph{``Imagine you hear about a movie that has taken in
10 million dollars at the box office, but don’t know how long it has been
running. What would you predict for the total amount of box office intake for
that movie?  The answer must be a single number, not a range, with no
explanation.''.  } I manually went through the results and extracted the
response value when applicable, rejecting answers where a range or a
nonsensical value was given (the number of rejected answers is presented in~\autoref{tab:pctnan}).
The Model answers and relevant code for reproducing the results are made
available online~\footnote{\url{https://github.com/sotlampr/llm-cognitive-judgements}}

\begin{table}[ht!]
\centering
\begin{tabular}{lrrr}
\toprule
                 &   Bayesian Prediction &   GPT-3 &   ChatGPT \\
\midrule
 Cakes           &                 11.5 &    30.1 &      16.5 \\  
 Life Spans      &                   2.5 &     5.6 &       3.4 \\ 
 Movie Grosses   &                  20.2 &    34.9 &     160.5 \\ 
 Poems           &                  5.6 &    28.6 &      20.1 \\  
 Representatives &                 16.0 &    32.0 &      29.5 \\  
 Waiting Times    &                 20.7 &    31.3 &      37.3 \\ 
\bottomrule
\end{tabular}
  \caption{Mean Average Percentage Error between the medians of the proposed models and the human judgements}\label{tab:mape}
\end{table}

\section{Results}
I present the Mean Average Percentage Error between each model and the human
judgements in~\autoref{tab:mape}. The Bayesian predictors
of~\cite{griffiths2006optimal} have consistently the lowest error compared to
either GPT-3 or ChatGPT. These results suggest that LLMs do not align with
human cognition when it comes to inductive cognitive judgments about everyday
phenomena under limited available data.

Comparing GPT-3 to ChatGPT, the latter is closer to human participants for most
tasks, with the striking exception of the ``Movie Grosses'', where ChatGPT was
the most hesitant to answer.

\begin{figure}
\centering
\input{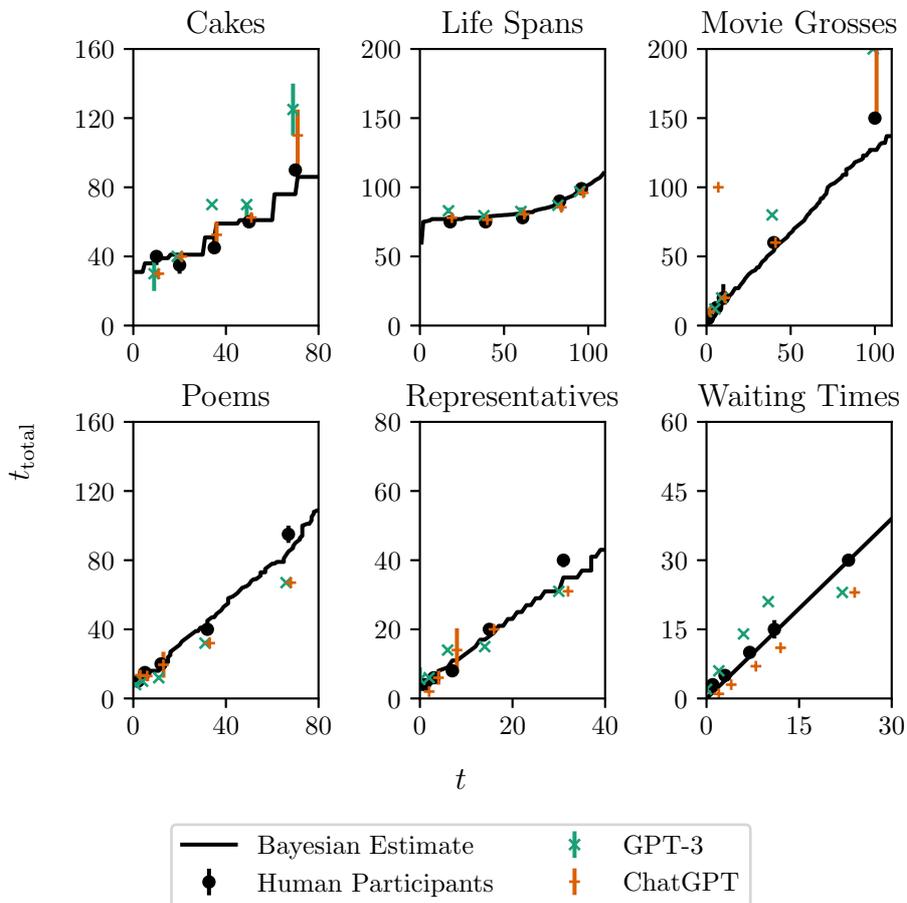}
  \caption{Medians and bootstrap 68\% confidence interval ($n=1000$)}\label{fig:plot}
\end{figure}

Graphs similar to~\cite{griffiths2006optimal} can be seen in ~\autoref{fig:plot}.
GPT-3 and ChatGPT demonstrate inconsistent predictions in a few cases regardless
of the task, as seen in the empirical confidence intervals for e.g. ``Cakes''
at $t=70$, ``Movie Grosses'' at $t=100$ and ``Representatives'' at $t=7$.

\begin{table}[ht!]
\centering
\begin{tabular}{lrr}
\toprule
 Task            &   GPT-3 &   ChatGPT \\
\midrule
 Cakes            &     0.0 &       0.0 \\
 Life Spans        &     0.0 &      45.0 \\
 Movie Grosses     &     0.0 &      71.3 \\
 Poems            &     0.0 &       0.0 \\
 Representatives  &     0.0 &       3.0 \\
 Waiting Times     &     0.0 &       0.0 \\
\bottomrule
\end{tabular}
  \caption{Percentage of samples with no satisfactory answer}\label{tab:pctnan}
\end{table}

I present the percentage of instances where no meaningful response was given
in~\autoref{tab:pctnan}. GPT-3 was able to answer meaningfully and succinctly in all cases,
while ChatGPT was hesitant when it comes to ``Life Spans'' and ``Movie Grosses''.
Common responses for both tasks were similar to
\emph{``The answer is not possible to determine$\dots$''}, 
\emph{``There is not enough information given to accurately predict$\dots$''},
\emph{``I'm sorry, but I cannot provide $\dots$''},
\emph{``Since I am an AI language model, I cannot$\dots$''}.
Additionally, on the ``Life Spans'' tasks, ChatGPT answered 6 times that it would
not be ethical to make such predictions.

These patterns in refusing to answer also appear in previous works such 
as~\cite{borji2023categorical,taecharungroj2023can,lloyd2023like,michaux2023can},
possibly revealing the effect of prompt tuning done with InstructGPT~\cite{ouyang2022training},
penalizing hallucinatory, counterfactual, or otherwise undesirable responses

\section{Conclusion}
I demonstrated a clear failure of popular LLMs when it comes to making
inductive judgements with limited available data on everyday scenarios. Despite
having a huge number of parameters and being trained on a huge amount of data,
these models cannot accurately model basic statistical principles that the
human mind seems to entrust, and which could be more accurately modelled with
many orders of magnitude less parameters.

% \section*{Acknowledgement}
% I would like to thank Agata Wlaszczyk for her input on the first draft.

% \clearpage

\bibliography{bibliography}
\end{document}